\documentclass{article}

\usepackage{microtype}
\usepackage{graphicx}
\usepackage{subfigure}
\usepackage{booktabs} 

\usepackage{hyperref}

\usepackage[preprint]{icml2023}

\usepackage{amsmath}
\usepackage{amssymb}
\usepackage{mathtools}
\usepackage{amsthm}

\usepackage{pifont}
\usepackage{diagbox} 
\usepackage{multirow}
\usepackage{caption}
\usepackage[capitalize,noabbrev]{cleveref}
\usepackage{colortbl}
\definecolor{mygray}{gray}{.9}

\theoremstyle{plain}
\newtheorem{theorem}{Theorem}

\theoremstyle{definition}
\newtheorem{definition}[theorem]{Definition}

\theoremstyle{remark}

\renewcommand{\algorithmicrequire}{ \textbf{Input:}}     
\renewcommand{\algorithmicensure}{ \textbf{Output:}}    

\usepackage[textsize=tiny]{todonotes}

\icmltitlerunning{Federated Open-world Semi-supervised Learning}

\begin{document}

\twocolumn[
\icmltitle{Towards Unbiased Training in Federated Open-world Semi-supervised Learning}



\icmlsetsymbol{equal}{*}

\begin{icmlauthorlist}
\icmlauthor{Jie Zhang}{yyy}
\icmlauthor{Xiaosong Ma}{yyy}
\icmlauthor{Song Guo}{yyy}
\icmlauthor{Wenchao Xu}{yyy}
\end{icmlauthorlist}

\icmlaffiliation{yyy}{Department of Computing, The Hong Kong Polytechnic University, Hong Kong, China}

\icmlcorrespondingauthor{Song Guo}{song.guo@polyu.edu.hk}

\icmlkeywords{Machine Learning, ICML}

\vskip 0.3in
]



\printAffiliationsAndNotice{}  

\begin{abstract}
   Federated Semi-supervised Learning (FedSSL) has emerged as a new paradigm for allowing distributed clients to collaboratively train a machine learning model over scarce labeled data and abundant unlabeled data.
   %
   However, existing works for FedSSL rely on a closed-world assumption that all local training data and global testing data are from seen classes observed in the labeled dataset.
   It is crucial to go one step further: \textit{adapting FL models to an open-world setting, where unseen classes exist in the unlabeled data.}
   %
   In this paper, we 
   propose a novel \textbf{\underline{Fed}}erated \textbf{\underline{o}}pen-world \textbf{\underline{S}}emi-\textbf{\underline{S}}upervised \textbf{\underline{L}}earning (\textbf{FedoSSL}) framework, which 
   can solve the key challenges in distributed and open-world settings, i.e., the biased training process for heterogeneously distributed unseen classes. Specifically, since the advent of a certain unseen class depends on a client basis, the locally unseen classes (exist in multiple clients) are likely to receive differentiated superior aggregation effects than the globally unseen classes (exist only in one client). 
   We adopt an uncertainty-aware suppressed loss to alleviate the biased training between locally unseen and globally unseen classes. Besides, we enable a calibration module supplementary to the global aggregation to avoid potential conflicting knowledge transfer caused by inconsistent data distribution among different clients. The proposed FedoSSL can be easily adapted to state-of-the-art FL methods, which is also validated via extensive experiments on benchmarks and real-world datasets (CIFAR-10, CIFAR-100 and CINIC-10).
\end{abstract}


\begin{figure*}[t]
\centering
    \includegraphics[width=0.95\textwidth]{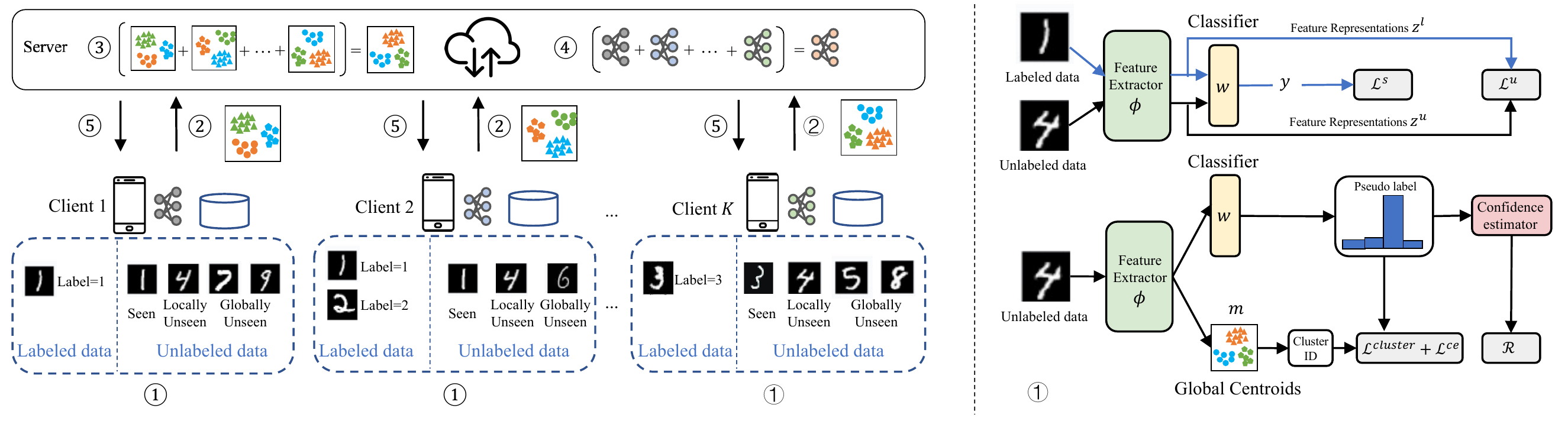}
    \caption{Framework of the proposed FedoSSL algorithm. \textbf{Pipeline}: \ding{172} 
    \textbf{Local Training:} Each client first performs local training on its private dataset for several epochs (i.e., via optimizing loss function in Eq.~(\ref{eq:L_i})), and then computes local centroids via a Sinkhorn-Knopp based clustering algorithm~\cite{genevay2019differentiable}. \ding{173} Upload model parameters and local centroids to the server. \ding{174} The server performs standard model aggregation. \ding{175} The server performs centroids aggregation by again using Sinkhorn-Knopp clustering to obtain global centroids. \ding{176} The global model and global centroids are returned to the clients, who use them for local training.}
    \label{fig:framework}
\end{figure*}

\section{Introduction}
\label{sec:intro}


To tackle the privacy issues in distributed machine learning, Federated Learning (FL) 
\cite{mcmahan2017communication,zhang2021edge} 
has emerged as a promising paradigm by collaboratively training a shared model among multiple clients without exposing their private raw data. 
As a common practice in FL, the global model is usually obtained by periodically averaging the updated model parameters from distributed clients in a centralized server. 
While existing FL methods assume that clients'
data is fully labeled so that supervised learning can be conducted for local model update on each client, in some real-world applications, the data labeling process can be prohibitive due to the tremendous overhead and the requirement of corresponding expertise \cite{ouali2020overview} such as for medical diagnoses \cite{ng2021federated} and object detection \cite{liu2021federated}. 
%

The scarce labeled data and the abundant unlabeled data give the rise to the emergence of federated semi-supervised learning (FedSSL) \cite{jeong2021federated,liang2022rscfed}, which can simultaneously exploit both the labeled and unlabeled data to optimize a global model in distributed environments. Existing FedSSL schemes have demonstrated to train models based on a small amount of labeled data on both client or server side \cite{jin2020towards,long2020fedsiam}. However, these works rely on the closed-world assumption that all local training data and global testing data are from the same set of classes that are included in the labeled dataset \cite{zhou2022open,boult2019learning}, which is often invalid for practical scenarios. In contrast, the open-world settings allow novel class discovery, and thus are common for data in-the-wild, e.g.,  in medical image classification tasks, some diseases are naturally scarce and never labeled before, while the model may be required to both classify images into predefined types (seen classes) and discover new unknown diseases (unseen classes). Hence,  a new fundamental question arises: \textit{how to collaboratively train models on distributed data to enable classification on both seen and unseen classes under the open-world setting?}

To this end, we first construct a new FedSSL benchmark that extends the conventional closed-world training framework to the open-world setting. Surprisingly, it is demonstrated that there is a significant performance degradation due to the existence of unseen classes in unlabeled data during the training process.
Although a few works explore the unseen class problem to avoid misclassifying unlabeled examples from unseen classes into seen classes, meanwhile alleviating the undesired performance gap between seen and unseen classes \cite{cao2022openworld,guo2022robust}, there is no literature considers this problem in a distributed environment.
With multiple participants, the problem definition is different, i.e., some unseen classes in one client may exist in other clients' side from a global view, and thus requires a novel fine-grained definition on unseen classes as well as the training mechanism for different types of samples/classes. It is worth mentioning that due to the heterogeneous distributed classes across different clients, simply aggregating the parameters following traditional FL mechanism can cause the biased training process for clients possessing different unseen classes.      


To tackle the above challenges, in this paper, we propose a brand-new \textbf{\underline{Fed}}erated \textbf{\underline{o}}pen-world \textbf{\underline{S}}emi-\textbf{\underline{S}}upervised \textbf{\underline{L}}earning framework (FedoSSL) that can achieve unbiased training procedure among different types of samples.
Specifically, the unseen classes in local clients are first redefined into \textit{locally unseen classes} and \textit{globally unseen classes}, respectively.
Then, we design a uncertainty-aware suppressed loss to adaptively control the undesired training divergence between locally and globally unseen classes, i.e. locally unseen classes often have higher training efficiency than globally unseen ones due to the cross-client collaboration. 
%
Furthermore, considering the heterogeneously distributed unseen classes across different clients that lead to potential conflicting knowledge transfer during the aggregation phase, we introduce a calibration module to produce corresponding global centroids to perform dedicated local adjusting on each client.  
%
%
We show that our method can significantly improve the model accuracy in a open-world setting when compared with the state-of-the-art baselines over widely used models (i.e., ResNet-18) and downstream tasks (i.e., CIFAR-10, CIFAR-100, CINIC-10).
%
The contributions of the paper are summarized as follows.


\begin{itemize}
    \item 
    To the best of our knowledge, we are the first to consider the open-world setting in FedSSL, where unseen classes exist in the unlabeled data, which is challenging due to the heterogeneously distributed unseen classes.
    %
    \item 
    We design a brand-new FedoSSL framework, that can achieve unbiased learning among different types of classes (i.e., locally unseen and globally unseen classes) and calibrated knowledge aggregation given heterogeneous data distributions.
    \item 
    We conduct extensive experiments on three typical image classification tasks. The empirical evaluation shows the superior performance of FedoSSL over the state-of-the-art approaches.
\end{itemize}

\section{Related Work}\label{sec:related_work}

\begin{table*}[t]
\setlength{\abovecaptionskip}{2pt}
\caption{Comparison between our proposed FedoSSL and other SSL-related methods.}
\label{tab:comparision}
\begin{center}
\begin{small}
\begin{tabular}{lccccc}
\toprule
  & \multicolumn{2}{c}{Training Dataset} & \multicolumn{2}{c}{Testing Dataset} &  \\
Method & Seen classes & Unseen classes & Seen classes & Unseen classes & FL Environment? \\
\midrule
SSL    & Present & Not Present & Classify & - & $\times$ \\
Open-set SSL & Present & Present & Classify & Detect \& Reject & $\times$\\
Novel Class Discovery    & Present & Present & - & Discover \& Cluster & $\times$ \\
Open-world SSL   & Present & Present & Classify &  Discover \& Cluster  & $\times$     \\
\midrule
FedSSL     & Present & Not Present & Classify  & - & $\surd$\\
FedoSSL     & Present & Present & Classify  & Discover \& Cluster & $\surd$\\
\bottomrule
\end{tabular}
\end{small}
\end{center}
\end{table*}

\subsection{Federated Learning}
Federated Learning (FL) has emerged as a promising paradigm to collaboratively train machine learning models using decentralized training data with privacy protection. As a most popular supervised FL training framework, FedAvg \cite{mcmahan2017communication} periodically aggregates the local updates at the server and transmits the averaged model back to local clients for the next round of training. In FL, Non-IID data is one of the key challenges due to the caused weight divergence and performance drop \cite{Li2020}. Many solutions have been proposed to solve this problem, i.e., sharing a public dataset among clients to achieve an approximately IID data \cite{Zhao2018,Yoshida2019}, data augmentation \cite{Jeong2018}, knowledge distillation \cite{Li2019,LinKSJ20}, regularization-based methods \cite{DBLP:conf/nips/0001MO20,li2020federated} and even personalized FL \cite{collins2021exploiting,zhang2021parameterized}. However, FL in the semi-supervised setting is less explored. 

\subsection{Semi-supervised Learning}
Semi-Supervised Learning (SSL) refers to the general problem of learning with partially labeled data, especially when the amount of labeled data is much smaller than that of the unlabeled data~\cite{zhou2005tri,rasmus2015semi}. The mainstream of SSL can be divided into two types:  pseudo-labeling and consistency regularization.
Pseudo-labeling methods~\cite{lee2013pseudo,pham2021meta,zou2020pseudoseg} usually generate artificial labels of unlabeled data from the model trained on labeled data and apply the filtered high-confidence labels as supervised signals for unlabeled data training. Consistency regularization methods~\cite{tarvainen2017mean,wei2020theoretical,lee2022contrastive} focus on training models via minimizing the distance among different perturbed outputs of the same input. 
All the above literature assume a closed-world setting in which labeled and unlabeled data come from the same set of predefined classes. However, this assumption rarely holds for data in real-world applications, namely, unseen classes would occur in unlabeled data. There are two recent lines of works related to different scenarios:   

\textbf{Open-set Semi-supervised Learning.} Open-set SSL considers that unseen classes in unlabeled samples only exist in training data, while not exist in testing data~\cite{chen2020semi,guo2020safe,huang2021trash,saito2021openmatch}.
The goal is to detect these unseen classes and reject them to ensure no performance degradation on seen classes. 

\textbf{Novel Class Discovery (NCD).} Different from Open-set SSL that fails to classify unseen classes, NCD~\cite{han2019learning,hsu2019multi,zhong2021neighborhood} aims to classify both seen and unseen classes during the testing phase but assumes all unlabeled instances belonging to unseen classes in training data. 

To extend unseen classes into a more practical scenario, open-world SSL~\cite{cao2022openworld,guo2022robust} focus on solving the class mismatch between labeled and unlabeled data, where each test sample should be either classified into one of existing classes or a new unseen class in the test time.

\subsection{Federated Semi-supervised Learning}
Learning representations from unlabeled decentralized data while preserving data privacy is still a nascent field. Existing FedSSL frameworks can be categorized into two types according to the location of the labeled data: 1) Labels-at-Server assumes that clients have purely unlabeled data and the server contains a small amount of labeled data~\cite{lin2021semifed,he2021ssfl,zhang2021improving,diao2022semifl}; 2) Labels-at-Client considers labeled data are available at local clients~\cite{jeong2021federated,lin2021semifed,liang2022rscfed}, which can be further subdivided into two scenarios: a) each client contains both labeled and unlabeled data~\cite{jeong2021federated}; b) some clients are fully labeled while some clients only contain unlabeled samples~\cite{lin2021semifed,liang2022rscfed,liu2021federated}. In this paper, we focus on the type 2.a as it has been largely overlooked and is more general in real-world FL scenarios.

\subsection{Summary}
To sum up, considering the existence of unseen classes in FedSSL is still a vacancy. More importantly, the characteristic of data heterogeneity among multiple clients brings new challenges for the combination of open-world and conventional FedSSL frameworks: \textit{how to tackle heterogeneous unseen classes among the clients in an efficient way?} Inspired by the above observations, we are motivated to develop a novel federated open-world semi-supervised learning framework to achieve high performance on both seen and unseen classes. The detailed comparisons between or proposed FedoSSL and other SSL-related methods are illustrated in Table~\ref{tab:comparision}.

\section{Methodology}\label{sec:methodology}
In this section, we first define the problem and 
elaborate the proposed FedoSSL framework, which contains an uncertainty-aware suppressed loss for local training and a calibration module for model aggregation. Then, we present the optimization process in Section~\ref{subsec:alg}.

\subsection{Preliminary and Problem Definition}
We focus on federated semi-supervised learning in an open-world setting, where the data in each client is partially labeled and the set of classes/categories in labeled data and unlabeled data are not the same, i.e., unseen classes exist in unlabeled data.
%
Assuming $K$ clients that each client $i$ holds a private classification dataset containing a labeled part $\mathcal{D}_i^l=\{(x_j, y_j) \}_{j=1}^{n_i^l}$ and an unlabeled part $\mathcal{D}_i^u=\{(x_j)\}_{j=1}^{n_i^u}$, where $n_i^l \ll n_i^u$, $x \in \mathbb{R}^{N}$, $1 \leq i \leq K$. 
The whole label and unlabeled dataset can be represented as $\mathcal{D}^l = \{\mathcal{D}_i^l\}_{i=1}^K$ and $\mathcal{D}^u = \{\mathcal{D}_i^u\}_{i=1}^K$, respectively. 
%
We denote the set of classes seen in the full labeled data as $\mathcal{C}^l$ and the set of classes in the unlabeled test data as $\mathcal{C}^u$. 
Unlike traditional (closed-world) FedSSL that $\mathcal{C}^l=\mathcal{C}^u$, in this paper, we consider $\mathcal{C}^l \neq \mathcal{C}^u$, and denote $\mathcal{C}_{seen}=\mathcal{C}^l \cap \mathcal{C}^u$, 
$\mathcal{C}_{unseen}=\mathcal{C}^u \setminus \mathcal{C}_{seen}$ 
as the set of seen classes and unseen classes, respectively.

The goal of FedSSL is to train a generalized global model $f$ with parameter $\theta$ from multiple decentralized clients, i.e., $\min_\theta \mathcal{L}(\theta) := \sum_{i=1}^K \frac{n_i}{n}\mathcal{L}_i(\theta)$, where $n_i=n_i^l+n_i^u$ and $n=\sum_{i=1}^K n_i$ is the total data amount. $\mathcal{L}_i(\theta)$  is the training loss function for client $i$. Specifically, the model $f$ can be decomposed of a feature extractor $g$ with parameter $\phi$: $\mathbb{R}^N \rightarrow \mathbb{R}^d$ to learn a low-dimensional feature $z$ and a classifier $h$ with parameter $w$: $\mathbb{R}^d \rightarrow \mathbb{R}^{|\mathcal{C}_{seen} \cup \mathcal{C}_{unseen}|}$.
%
%
%
The training loss of a semi-supervised learning algorithm on each client $i$ usually contains supervised loss $\mathcal{L}_i^s$ and unsupervised loss $\mathcal{L}_i^u$ with weight parameters $\alpha$, $\alpha >0$:
\begin{equation}
    \mathcal{L}_i = \mathcal{L}_i^s+\alpha\mathcal{L}_i^u
    \label{eq:L_i}
\end{equation}
%

Typically, $\mathcal{L}_i^s$ applies the standard cross-entropy loss on labeled instances:
\begin{equation}
    \mathcal{L}_i^s = \frac{1}{n_i^l} \sum_{(x_j, y_j) \in \mathcal{D}_i^l}  \mathcal{H}(y_j, p(x_j;\theta))
    \label{eq:L_is}
\end{equation}
where $p(x;\theta)=\text{Softmax}(f(x;\theta))$ denotes the predicted probabilities produced by the model $f$ for input $x$, and $\mathcal{H}(\cdot,\cdot)$ is the cross-entropy function. 

In terms of unsupervised loss $\mathcal{L}_i^u$, there are two typical forms: \textbf{pseudo-labels}~\cite{sohn2020fixmatch} based on labeled data and \textbf{consistency regularization}~\cite{xie2020unsupervised} based on data augmentation. However, in an open-world setting, the existence of unseen classes makes above methods fail to classify seen classes and unseen classes. Therefore, similar to ORCA~\cite{cao2022openworld} and NACH~\cite{guo2022robust}, we use pairwise objective as unsupervised loss on unlabeled data to classify unseen classes:
\begin{equation}
    \mathcal{L}_i^u = -\frac{1}{n_i^l+n_i^u} \hspace{-1.5em}
    \sum_{\substack{z_j,\bar{z}_j\in\\(Z_i^l \cup Z_i^u, \bar{Z}_i^l \cup \bar{Z}_i^u)}} 
    \hspace{-1.5em}
    \mathcal{H}(p(w^{\top}\cdot z_j), p(w^{\top}\cdot \bar{z}_j))
    \label{eq:L_iu}
\end{equation}
where $Z_i^l$ and $Z_i^u$ are the whole set of feature representations for labeled and unlabeled data, respectively.
$\bar{Z}_i^l \cup \bar{Z}_i^u$ is the closet set of $Z_i^l \cup Z_i^u$ in a mini-batch by computing the cosine distance between all pairs of feature representations.

\subsection{Overview of FedoSSL}
Previous FedSSL methods do not consider the existence of unseen classes on each client, which leads to a number of data from unseen classes being misclassified into seen classes. Besides, inconsistent data distribution on different clients raises another new problem: 
\textit{some unseen classes may exist in more than one client, resulting in biased training among different unseen classes},
e.g., in Figure~\ref{fig:framework}, both client 1, 2 and $K$ have class 4, while class 5, 6, 7, 8 and 9 only exist in one of the clients. 
In this case, we need a more fine-grained definition on unseen class in a global view.

\begin{definition}[locally unseen \& globally unseen class]
        In FedoSSL, the unseen classes $\mathcal{C}_{i, unseen}$ on client $i$ can be divided into two types: locally unseen classes $\mathcal{C}_{i, lu}$, in which $\mathcal{C}_{i, lu} = \mathcal{C}_{1, unseen} \cap \cdots \cap \mathcal{C}_{K, unseen}$; 
        and globally unseen classes $\mathcal{C}_{i, gu}$, in which $\mathcal{C}_{i, gu}=\mathcal{C}_{i, unseen} \setminus \mathcal{C}_{i, lu}$.
\end{definition}

When unseen classes exist among multiple clients in FedoSSL, two main challenges need to be considered. 
Firstly, locally unseen classes may be learned faster than globally unseen classes due to the facilitation of client collaboration on locally unseen classes. The existing unsupervised pairwise loss $\mathcal{L}_i^u$ treats each class equally, while the imbalanced training progresses among unseen classes will result in a big bias on pseudo-label generation and even performance degradation in seen classes. Thus, we propose an uncertainty-aware regularization loss to alleviate training bias among different classes.
Moreover, feature-level unsupervised pairwise loss on unlabeled data makes the generated cluster/class id heterogeneous among different clients since both labeled data and unlabeled data are required to feed into the same model classifier. It is critical to design a calibration strategy to align the outputs of the same unseen classes during the model aggregation phase. 

\subsubsection{Objective}
To achieve unbiased training among different unseen classes in FedSSL, we present the proposed FedoSSL method. The overall objective consists of three parts: 1) fundamental semi-supervised loss for all data; 2) an uncertainty-aware regularization loss to reduce the training gap among locally unseen and globally unseen classes; 3) a calibration loss to achieve efficient model aggregation:
\begin{equation}
    \mathcal{L}^{*}_i  = \mathcal{L}_i + \beta \mathcal{R}_i 
    + \gamma \mathcal{L}_i^{cal}
    \label{eq:client_objecitve}
\end{equation}
where $\beta$ and $\gamma$ are trade-off hyper-parameters.

\subsubsection{Uncertainty-aware Loss}
Considering that different types of unseen classes have various training progress, i.e, locally unseen classes can be facilitated from model collaboration, we seek to add a regularization term to alleviate training divergence between locally unseen and globally unseen classes. Specifically, we use the uncertainty of the generated cluster/class id to reflect the training progress and apply a larger penalty for samples with high uncertainty. Then, the uncertainty-aware loss can be defined as:
\begin{equation}
    \mathcal{R}_i = \frac{1}{n_i^u}\sum_{x_j^u \in \mathcal{D}_i^u} |\pi(x_j^u)|
\end{equation}
where $\pi(\cdot)$ is the data uncertainty function. To precisely explore the uncertainty, we rely on both the confidence of the pseudo-label computed from the output of the softmax function and the proportion of samples belonging to related pseudo-label, i.e., 
\begin{equation}
    \pi(x_j^u) = \rho(n^c|\arg\max_c p(x_j^u;\theta)) [1-\max_c p(x_j^u;\theta)] 
\end{equation}
where $\rho(n^c) = -\tau^{1-\frac{n^c}{n_{\text{max}}}}$ is the weight for class $c$, which can be estimated from the labeled data, $\tau \in (0,1]$. $n^c$ is the number of training samples of the class $c$ predicted by the model. $n_{\text{max}}$ is the number of samples of the class with the maximal size.
Note that $\rho(n^c)$ can be any function inversely proportional to $n^c$. In this paper, we focus more on unbiased training among locally unseen and globally unseen classes, while the training inconsistency between seen classes and unseen classes is ignored as 
both ORCA~\cite{cao2022openworld} and NACH~\cite{guo2022robust} have tackled this problem.

\subsubsection{Calibration Module}
\begin{figure}[t]
\centering
    \includegraphics[width=0.45\textwidth]{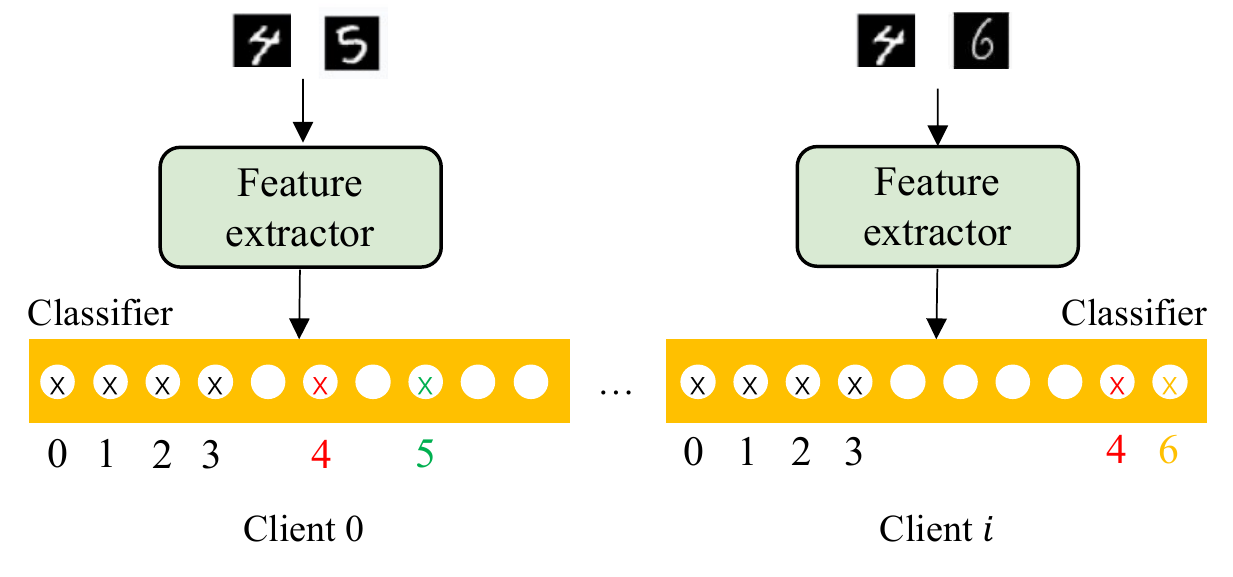}
    \caption{Illustration of label heterogeneity in FedoSSL. In a 10-class classification example, classes \{0, 1, 2, 3\} are seen classes, while classes \{4, 5, 6\} are unseen classes. Due to the feature-level pair-wise unsupervised loss (i.e., $\mathcal{L}_i^u$) on unlabeled data, same unseen class would be classified with different label id on different clients, e.g., unseen class 4 would be classified into the sixth position of the client 0's classifier, while in client $i$ class 4 would be classified into the eighth position of the classifier.     
    }
    \label{fig:label_hetero}
\end{figure}

Due to the unseen classes being classified into new clusters by the pair-wise objective, i.e., $\mathcal{L}_i^u$, which only ensures similar samples are classified into one group/cluster, the same unseen class may be classified with different cluster id on different clients. 
For example, in Figure~\ref{fig:label_hetero}, unseen classes 4, 5 and 6 would be classified with any labels between 4$\sim$9 because of lacking supervision from the labeled data. Such label heterogeneity would significantly degrade the aggregation performance. Therefore, it is required to design a calibration module to align the heterogeneous outputs of local classifiers before the aggregation phase.  

Inspired by a clustering-based FL technique~\cite{lubana2022orchestra} that aims to align local clustering performances among different clients via adding a global centroids aggregation mechanism, we extend this technique to our FedoSSL scenario and use the global centroids as a self-supervised signal to guide the classification process on unseen classes. The loss of the calibration module can be represented as
\begin{equation}
    \mathcal{L}_i^{cal} =  \mathcal{L}_i^{ce} +\mathcal{L}_i^{cluster}
\end{equation}

Specifically, we first aggregate local centroids from all clients to obtain global clusters at the server, i.e, by using Sinkhorn-Knopp~\cite{genevay2019differentiable} clustering. The global centroids are then returned to the clients for further calibration.
Taking the global centroids as guidance, the outputs of the classifier on unseen classes can be updated by approaching the cluster assignments with a cross-entropy loss:
\begin{equation}
    \mathcal{L}_i^{ce} = \frac{1}{n_i^u} \sum_{z_j \in Z_i^u} \mathcal{H}(q(z_j;m), p(w^{\top} \cdot z_j))
\end{equation}
where $m$ denotes the global centroids.
Moreover, to prevent the cluster assignment of one class from changing dramatically during the training process, we additionally design the following loss function for promoting clusterability of feature representations:
\begin{equation}
    \mathcal{L}_i^{cluster} = \frac{1}{n_i^u} \sum_{z_j, \bar{z}_j \in Z_i^u, \bar{Z}_i^u } \mathcal{H}(q(z_j;m), q(\bar{z}_j;m))
\end{equation}
where $q(z_j;m)$ and $q(\bar{z}_j;m)$ are corresponding cluster assignments, which are computed by matching representations with global centroids.

\subsection{Algorithm Workflow}\label{subsec:alg}
In this subsection, we illustrate the overall learning framework of FedoSSL, which is summarized in Algorithm~\ref{alg:1}. 
%


\textbf{Client Update.} In each communication round, the clients download the global model and global centroids from the server. In the client update phase, each of them makes several local gradient-based updates (e.g., $E$ epochs) to optimize the local objective in Eq.~(\ref{eq:client_objecitve}). Then, local centroids are computed via a Sinkhorn-Knopp based clustering algorithm~\cite{genevay2019differentiable}. 

\textbf{Server Aggregation.} After training the local models at the client-side, both updated models and local centroids will be sent to the server for further aggregation. 
Specifically, the server first aggregates the local models by taking a weighted average of them. Then, global centroids are calculated by aggregating local centroids (i.e., again using Sinkhorn-Knopp clustering). The above steps will be repeated until the model achieves convergence.

\begin{algorithm}[tb]
   \caption{FedoSSL Algorithm}
   \label{alg:1}
\begin{algorithmic}[1]
   \REQUIRE Number of clients $K$, learning rate $\eta$, local epochs $E$, labeled data $\{\mathcal{D}_1^l, \mathcal{D}_2^l,\dots,\mathcal{D}_K^l\}$, unlabeled data $\{\mathcal{D}_1^u, \mathcal{D}_2^u,\dots,\mathcal{D}_K^u\}$, hyperparameter $\alpha, \beta, \gamma$
   \ENSURE Final model $\theta$
   \STATE Initialize the model parameter $\theta$
   \REPEAT
   \STATE Sample a set of clients $\mathcal{S}$.
   \FOR {each client $i \in \mathcal{S}$ in parallel}
   \FOR{$j=1$ {\bfseries to} $E$}
   \STATE Update local model: $\theta_i \leftarrow \theta_i - \eta \nabla_{\theta}\mathcal{L}_i^{*}(\theta_i)$
   \ENDFOR
   \STATE Calculate local centroids $m_i$
   \ENDFOR
   \STATE Update global model: $\theta \leftarrow \frac{n_i}{\sum_{i \in \mathcal{S}} n_in}\sum_{i \in \mathcal{S}} \theta_i$ 
   \STATE Update global centroids $m$
   \STATE Distribute $\theta$ and $m$ to all clients
   \UNTIL{Model converges}
\end{algorithmic}
\end{algorithm}


\section{Experiments}

\subsection{Experimental Setup}
\textbf{Dataset.} We evaluate the FedoSSL framework over three datasets CIFAR-10, CIFAR-100, and CINIC-10~\cite{darlow2018cinic}. 
CINIC-10 is a larger dataset that is constructed from CIFAR-10 and ImageNet. 
For all datasets, we first divide classes into 60$\%$ seen and 40$\%$ unseen classes, then select 50$\%$ of seen classes as the labeled data and the rest as unlabeled data. For CIFAR-10 and CINIC-10 datasets, one class of unseen classes is selected as the globally unseen class and rest 3 classes are locally unseen classes, each client owns all 6 seen classes, one globally unseen class and one locally unseen class. For CIFAR-100 dataset, 10 classes of unseen classes are selected as the globally unseen class and rest 30 classes are locally unseen classes, each client owns all 60 seen classes, 10 globally unseen classes and 10 locally unseen classes.

\begin{table*}[t]
\centering
\small
\setlength{\tabcolsep}{1.4mm}{
\caption{Classification accuracy of compared methods on seen, unseen and all classes with 10 clients over three benchmark datasets. 
Asterisk (${\ast}$) in $^{\ast}$SemiFL denotes that the original methods cannot classify unseen classes (and we had to extend it).
On unseen classes, \textit{LU.} denotes locally unseen classes, while \textit{GU.} denotes globally unseen classes. \textit{AU.} represents the overall accuracy of all unseen classes. Gray rows indicate the upper bound of the model performance of FedoSSL.}
\begin{tabular}{lccccc|ccccc|ccccc}
\toprule
\multicolumn{1}{c}{ }  & \multicolumn{5}{c}{CIFAR-10 (\%)} & \multicolumn{5}{c}{CIFAR-100 (\%)} &
\multicolumn{5}{c}{CINIC-10 (\%)}   \\ 
\cmidrule(lr){2-6}\cmidrule(lr){7-11} \cmidrule(lr){12-16}
  \multirow{2}{*}{\#Method} &  \multirow{2}{*}{All} &  \multirow{2}{*}{Seen} & \multicolumn{3}{c|}{Unseen} & \multirow{2}{*}{All} & \multirow{2}{*}{Seen} & \multicolumn{3}{c|}{Unseen}  & \multirow{2}{*}{All} & \multirow{2}{*}{Seen} & \multicolumn{3}{c}{Unseen}    \\ 
  \cmidrule(lr){4-6}\cmidrule(lr){9-11}\cmidrule(lr){14-16}
  & & & LU.&GU. & AU.  & & &LU. & GU.& AU. & & & LU. & GU. & AU.\\
 \midrule
 \rowcolor{mygray} Cen-O &78.26 & 86.63 & -& -& 71.95& 56.92& 73.68& -& -& 44.28& 69.32& 83.18 & -&-& 58.86\\
 \rowcolor{mygray} Cen-N & 81.02& 89.47& -& -& 74.64& 58.98& 75.10& -& -& 46.82& 71.89& 83.82 & -&-&62.89\\
 \midrule
 Local-O & 65.98& 79.57& -& -& 45.60& 43.10& 54.33& -& -& 26.25& 55.33& 65.23 & -&-& 40.48\\
 Local-N & 67.67& 83.95& -& -& 43.26& 45.28& 57.24& -& -& 27.34& 57.31& 65.70 & -&-& 44.73\\
 \midrule
   Fed-AO & 69.46  & 81.01  & 89.38 &  42.03 & 52.15  &
 47.91 & 59.67  &  38.07 &  29.12 & 30.26  & 
 54.85 & 63.22   &  71.31 &  37.88 & 42.29 \\ 
    Fed-RO  & 71.72  & 82.22  & 89.84 & 53.43 & 55.96  &
 47.72 & 59.79  & 44.13 & 28.86 & 29.62  & 
 57.16 & 62.26   & 72.24 & 42.09 & 49.50 \\ 
    Fed-AN & 66.58  & 84.18  & 78.76 & 37.58 & 40.15  &
 47.25 & 58.24  & 42.11 & 30.44 & 30.77  & 
 53.49 & 63.61   &  66.78 & 36.06  &  38.32 \\ 
    Fed-RN  & 68.83  & \textbf{85.52}  & 79.84 & 41.79 & 43.81  &
 48.02 & 59.4  & \textbf{48.77} & 30.36 & 30.96  & 
 58.11 & 65.97   & 68.81 & 39.01  & 46.33 \\ 
  \midrule
   $^{\ast}$SemiFL  & 64.91	& 81.57 &	86.33	& 31.16	& 39.92	& 42.28 &	54.94	& 31.68	& 21.46	& 23.29	& 52.27	& 62.72	& 64.53	& 37.21	& 37.34 \\ 
\midrule
   \textbf{FedoSSL} & \textbf{76.26}  & 84.29 & \textbf{90.68} & \textbf{59.69}& \textbf{64.22}  &
 \textbf{51.58} & \textbf{61.12}  & 45.76 &  \textbf{33.82} & \textbf{31.13}  & 
 \textbf{63.82} & \textbf{68.40} &  \textbf{79.79}& \textbf{47.78} &\textbf{56.96}   \\ 
\bottomrule
 \label{tab:small_fl}
 \vspace{-0.5cm}
\end{tabular}}
\end{table*}

\textbf{Baselines.} To compare our FedoSSL with state-of-the-art methods, 
we conduct experiments from two perspectives: 1) extending existing open-world SSL methods to FL environments;
2) extending existing FedSSL methods to the open-world scenarios.
%

For the former one, we extend ORCA and NACH to be applicable to federated learning scenarios by implementing two representative FL algorithms with them, i.e., FedAvg~\cite{mcmahan2017communication} and FedRep~\cite{collins2021exploiting}. Specifically, the extended baselines are named FedAvg+ORCA (\textbf{Fed-AO}), FedRep+ORCA (\textbf{Fed-RO}), FedAvg+NACH (\textbf{Fed-AN}) and FedRep+NACH (\textbf{Fed-RO}). Note that FedRep is a personalized federated learning algorithm that keeps each client’s classifier updating locally, while the other parts are aggregated at the server. 
The reason for implementing FedRep is that the same unseen class may be classified with different labels on different classifiers due to the lack of uniform labels among different clients when extending ORCA and NACH to a distributed environment.
In order to avoid the confusion caused by the aggregation of local classifiers, FedRep does not aggregate the classifiers. 

For FedSSL methods, we adopt one state-of-the-art method \textbf{SemiFL}~\cite{diao2022semifl} to the open-world setting in the following way: we use SemiFL to classify samples into seen classes and estimate out-of-distribution (OOD) samples based on softmax confidence scores. Note that SemiFL has already shown superiority over FedMatch~\cite{jeong2021federated}, thus we ignore FedMatch as a baseline in the paper.

To ensure fairness in performance comparison, we also run the open-world SSL methods (i.e., ORCA, NACH) in the centralized (i.e., \textbf{Cen-O}, \textbf{Cen-N}) and distributed (i.e., \textbf{Local-O}, \textbf{Local-N}) environment without aggregation, respectively, to indicate the upper bound and lower bound of the performance of our FedoSSL method.


\textbf{Implementation Details.}
For all datasets, we use ResNet-18 as the backbone model and train the model using standard SGD with a momentum of 0.9 and a weight decay of $5 \times 10^{-4}$. 
The dimension of the classifier corresponds to the number of classes in each dataset.
Unless otherwise explicitly specified, $\alpha, \beta, \gamma$ are set to 1.
The model is trained for 50 global rounds with 5 local epochs in each round. The batch size is 512 for all experiments. 
Similar to ORCA~\cite{cao2022openworld}, we only update the parameters of the last block of ResNet in the second training stage to avoid over-fitting. 
For Sinkhorn-Knopp clustering~\cite{genevay2019differentiable} in our FedoSSL, we compute 32 local centroids and 10 global centroids for CIFAR-10 and CINIC-10 and 128 local centroids and 100 global centroids for CIFAR-100.
We evaluate all baselines and our FedoSSL in two settings: 10 clients with 50$\%$ participation ratio and 50 clients with 10$\%$ participation ratio. The final average model accuracy of all clients is obtained from the best round among all global rounds. 
All compared methods are implemented based on the pre-trained model using the contrastive learning algorithm  SimCLR~\cite{chen2020simple}.

We simulate all clients and the server on a workstation with an RTX 2080Ti GPU, a $3.6$-GHZ Intel Core i9-9900KF CPU and $64$GB of RAM.

\textbf{Remark.} We are aware that many tricks, e.g.,
a diversity-constrained regularization loss to avoid trivial solutions~\cite{cao2022openworld,guo2022robust} or a dynamic threshold to filter samples with lower confidence~\cite{guo2022robust}, etc., both can be added to our FedoSSL framework. However, to avoid losing focus, we ignore these tricks in this paper.

\subsection{Performance Comparison}
\begin{table}[tb]
\setlength{\abovecaptionskip}{2pt}
\caption{Analysis of Loss function: classification accuracy on CIFAR-10 (the number of clients: 10).}
\label{tab:analysis_on_loss_CIFAR10}
\begin{center}
\begin{small}
\begin{sc}
\begin{tabular}{lcccr}
\toprule
Method & Seen & Unseen & ALL \\
\midrule
Fed-AO    & 81.01 & 52.15 & 69.46 \\
FedoSSL-$\mathcal{R}_i$-$\mathcal{L}_i^{ce}$  & 83.53 & 52.24 & 71.01\\
FedoSSL-$\mathcal{R}_i$    & 83.13 & 62.98 & 75.07 \\
FedoSSL   & 84.29 & 64.22 &  76.26       \\
\bottomrule
\end{tabular}
\end{sc}
\end{small}
\end{center}
\vskip -0.18in
\end{table}

\begin{table}[tb]
\setlength{\abovecaptionskip}{2pt}
\caption{Analysis of Loss function: classification accuracy on CINIC-10 (the number of clients: 10). }
\label{tab:analysis_on_loss_CININ10}
\begin{center}
\begin{small}
\begin{sc}
\begin{tabular}{lcccr}
\toprule
Method & Seen & Unseen & ALL \\
\midrule
Fed-AO    & 63.22 & 42.29 & 54.85 \\
FedoSSL-$\mathcal{R}_i$-$\mathcal{L}_i^{ce}$  & 69.10 & 40.31 & 57.58\\
FedoSSL-$\mathcal{R}_i$    & 67.59 & 47.73 & 59.65 \\
FedoSSL   & 68.40 & 56.69 &  63.82       \\
\bottomrule
\end{tabular}
\end{sc}
\end{small}
\end{center}
\vskip -0.2in
\end{table}

\begin{figure*}[t]
\centering
\setlength{\abovecaptionskip}{0pt}
\subfigure[In initial stage.]{\label{subfig:a} 
 \includegraphics[width = 0.31\linewidth]{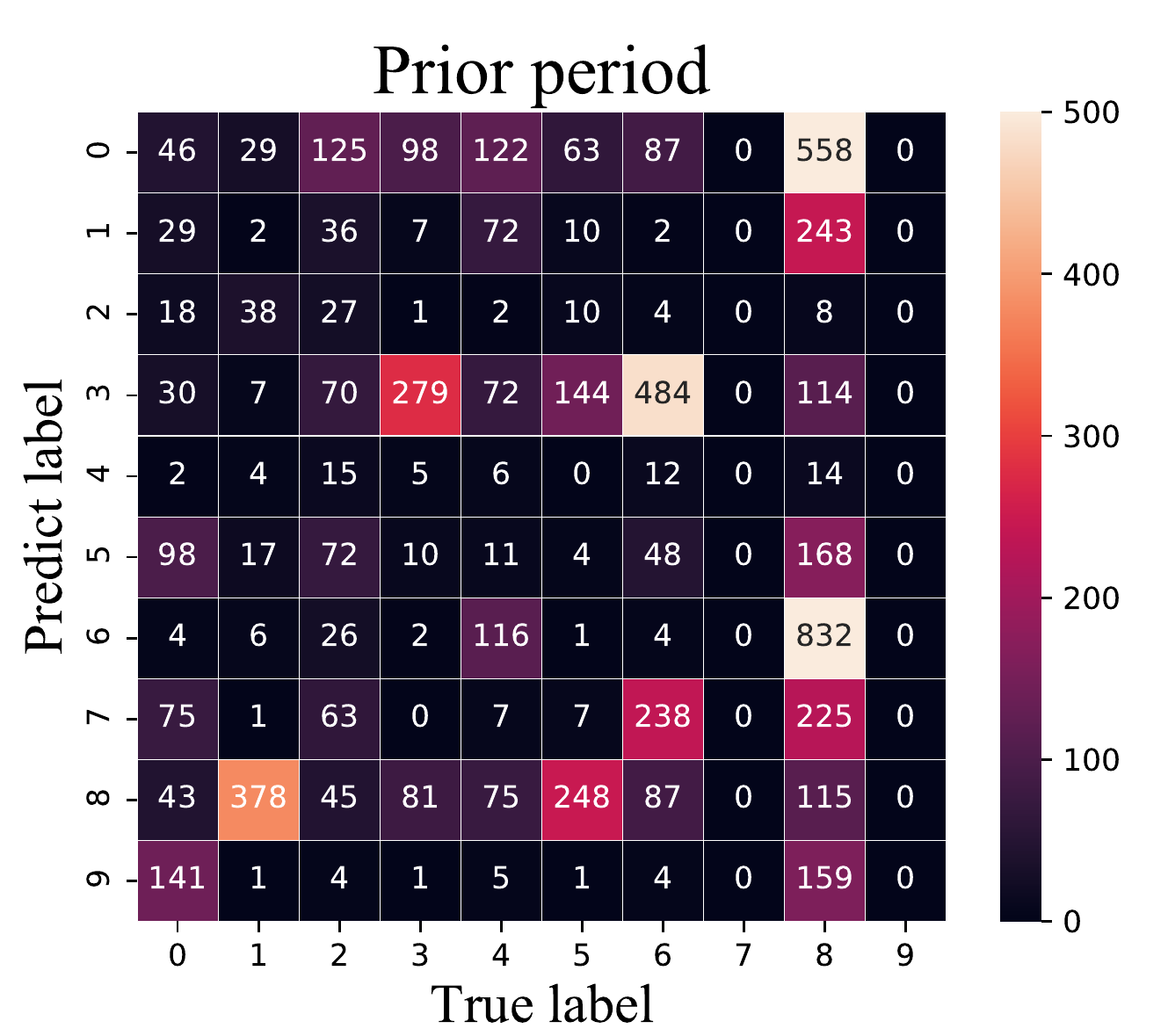}}
\subfigure[In intermediate stage.]{\label{subfig:b} 
 \includegraphics[width = 0.3\linewidth]{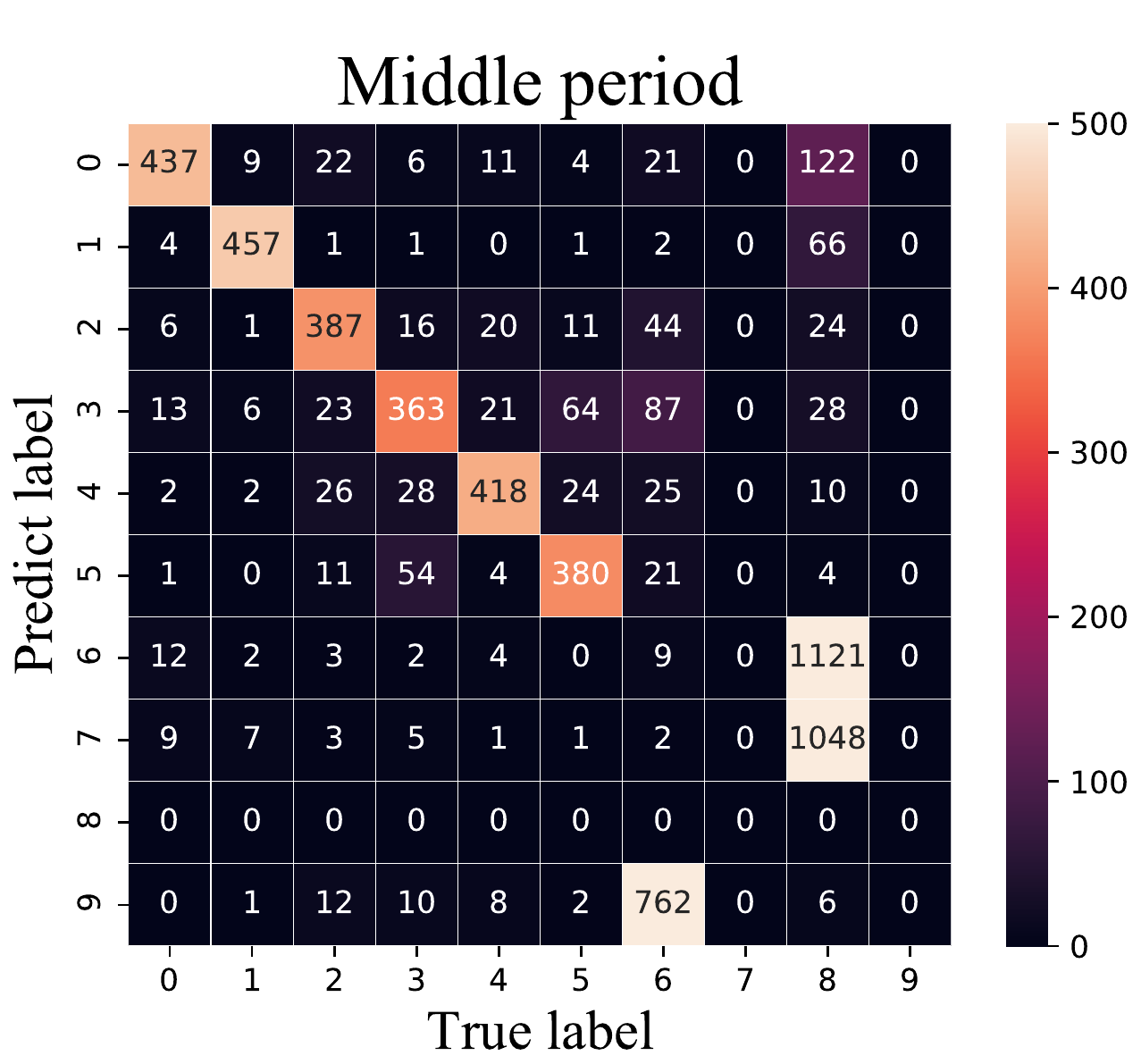}}
 \subfigure[In final stage.]{\label{subfig:c} 
 \includegraphics[width = 0.3\linewidth]{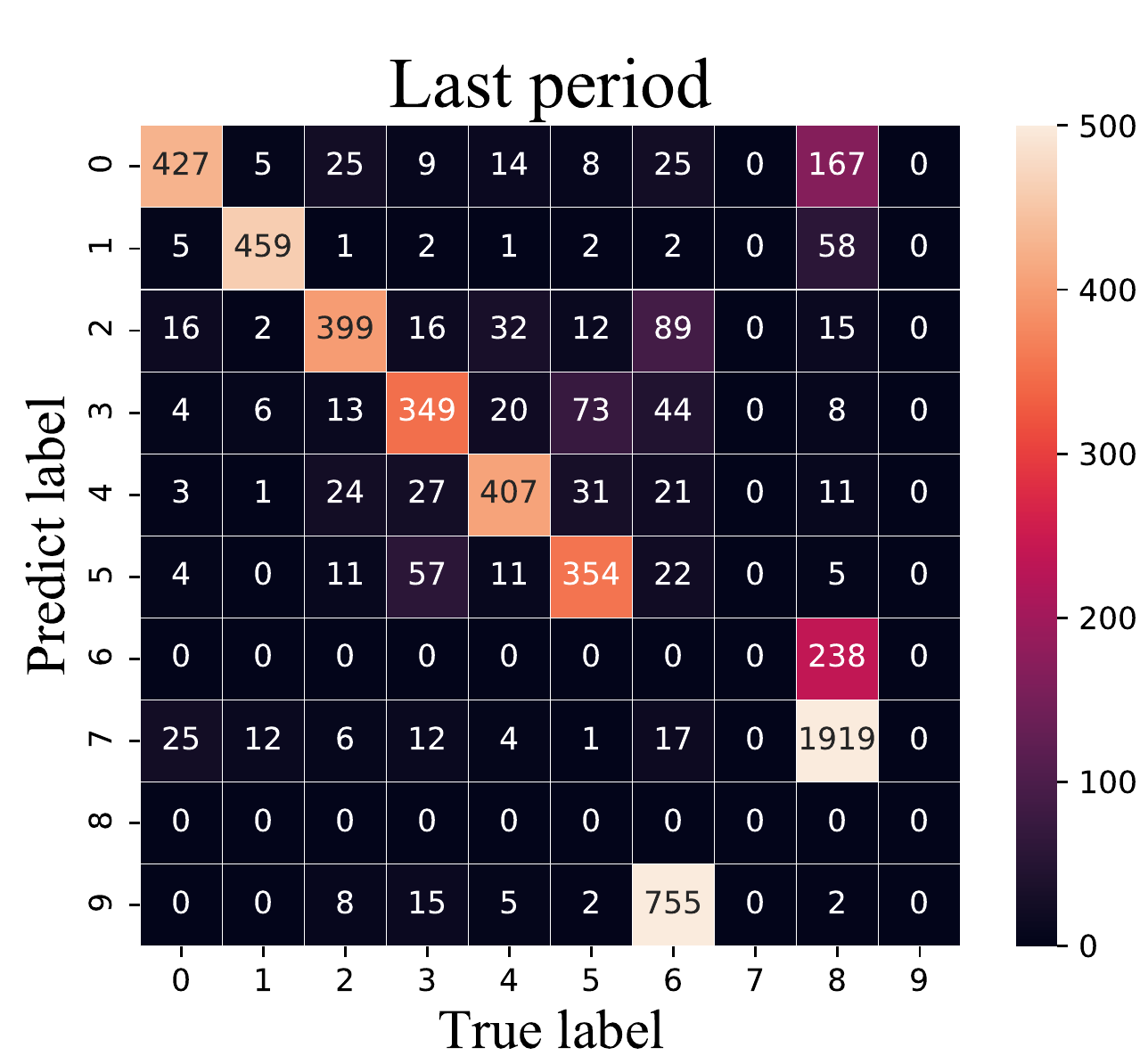}}
\caption{Classification accuracy on different 
 setting of $\beta$ (a) and $\gamma$ (b).}
\label{fig:visulizations}
\vspace{-0.25cm}
\end{figure*}

\begin{table*}[t]
\centering
\small
\setlength{\abovecaptionskip}{0pt}
\caption{Classification accuracy of compared methods on seen, unseen and all classes with 50 clients over three benchmark datasets. 
}
\begin{tabular}{lccc|ccc|ccc}
\toprule
\multicolumn{1}{c}{ }  & \multicolumn{3}{c}{CIFAR-10 (\%)} & \multicolumn{3}{c}{CIFAR-100 (\%)} &
\multicolumn{3}{c}{CINIC-10 (\%)}   \\ 
\cmidrule(lr){2-4}\cmidrule(lr){5-7} \cmidrule(lr){8-10}
   \#Method &  All &  Seen & Unseen & All & Seen & Unseen  & All & Seen & Unseen \\
 \midrule
   Fed-AO & 70.22  & 83.34   & 50.54  &
 45.63 & 56.25   & 29.69  & 
 53.81 & 60.49   & 43.80 \\ 
    Fed-RO  & 71.36  & 84.31   & 51.93  &
 45.18 & 56.78   & 27.79  & 
 57.26 & 61.70    & 50.61 \\ 
    Fed-AN & 69.89  & 85.36  & 46.68  &
 45.22 & 56.30   & 28.59  & 
 53.42 & 63.62   &  38.13 \\ 
    Fed-RN  & 71.49  & \textbf{86.28}  & 49.30  &
 45.57 & 56.79  & 28.73  & 
 57.81 & 65.29    & 46.60 \\ 
\midrule
   FedoSSL & \textbf{76.41}  & 85.71 & \textbf{62.46}  &
 \textbf{47.01} & \textbf{58.34}  & \textbf{30.17}  & 
 \textbf{64.02} & \textbf{69.56} &\textbf{55.71}   \\ 
\bottomrule
 \label{tab:large_scale_fl}
 \vspace{-0.6cm}
\end{tabular}
\end{table*}

First, we compare our proposed FedoSSL methods with the baseline methods over three benchmark datasets.
The classification accuracy on CIFAR-10, CIFAR-100 and CININ-10 dataset are listed in Table~\ref{tab:small_fl}.
It should be noted that the overall unseen accuracy (AU.) is not always a weighted average of locally unseen accuracy (LU.) and globally unseen accuracy (GU.). Because sometimes most samples of a locally unseen class and most samples of a globally unseen class will be classified in the same label, in this case, when calculating overall unseen accuracy, we have to choose another label (i.e., with the second largest number of samples) for one of those two classes.

From the results, it can be observed that our proposed FedoSSL provides superior performance of overall classification accuracy than baselines and the locally trained versions (i.e., Local-O, Local-N) over all three datasets. In most cases, FedoSSL maintains robust performance on seen classes and locally unseen classes. On globally unseen classes and overall unseen classes, FedoSSL achieves remarkable performance gains. For example, for globally unseen classes, FedoSSL outperforms the best baselines by 11.72\% on CIFAR-10, 11.10\% on CIFAR-100 and 13.52\% on CINIC-10. For overall unseen classes, FedoSSL outperforms the best baselines by 14.76\% on CIFAR-10 and 15.07\% on CINIC-10. 
Moreover, we prove that the performance gap between locally and globally unseen classes on FedoSSL has significantly reduced when compared with other methods, i.e., there is a 47\% gap between locally and globally unseen classes in Fed-AO, while FedoSSL reduces this gap to 31\% on CIFAR-10.

\subsection{Ablation Study}
In this subsection, detailed analyses are shown to help understand the superiority of our proposed FedoSSL framework, including analysis on the two additional modules: $\mathcal{R}_i$ and $\mathcal{L}_i^{cal}$, effect of the number of seen classes, robustness on large-scale of FL scenarios, and hyper-parameter sensitivity analysis.

\textbf{Analysis on Objective Functions.}
We evaluate our proposed two objective functions $\mathcal{R}_i$ and $\mathcal{L}_i^{cal}$ (i.e., $\mathcal{L}_i^{cal}$ consists of $\mathcal{L}_i^{ce}$ and $\mathcal{L}_i^{cluster}$) of FedoSSL on CIFAR-10 and CINIC-10 dataset. We use the Fed-AO as the basic baseline. First, FedoSSL-$\mathcal{R}_i$-$\mathcal{L}_i^{ce}$ means that only adding $\mathcal{L}_i^{cluster}$ to the baseline, it can be observed that the accuracy of seen classes has improved. Then, FedoSSL-$\mathcal{R}_i$ means that adding both $\mathcal{L}_i^{ce}$ and $\mathcal{L}_i^{cluster}$ could greatly improve the accuracy of unseen classes. Finally, the case of using all two objective functions, i.e., the complete FedoSSL, further improves the accuracy of the unseen classes and achieves the best performance. Results in Table~\ref{tab:analysis_on_loss_CIFAR10} and \ref{tab:analysis_on_loss_CININ10} give a clear ablation study to demonstrate the effectiveness of our proposed objective functions. 
%
Besides that, we record the calculated clustering assignments in the initial stage, intermediate stage and final stage, respectively. Figure~\ref{fig:visulizations} shows that sharing local centroids can significantly help to calibrate the heterogeneous outputs of the classifier on unseen classes.

\begin{figure}[t]
\centering
    \includegraphics[width=0.45\textwidth]{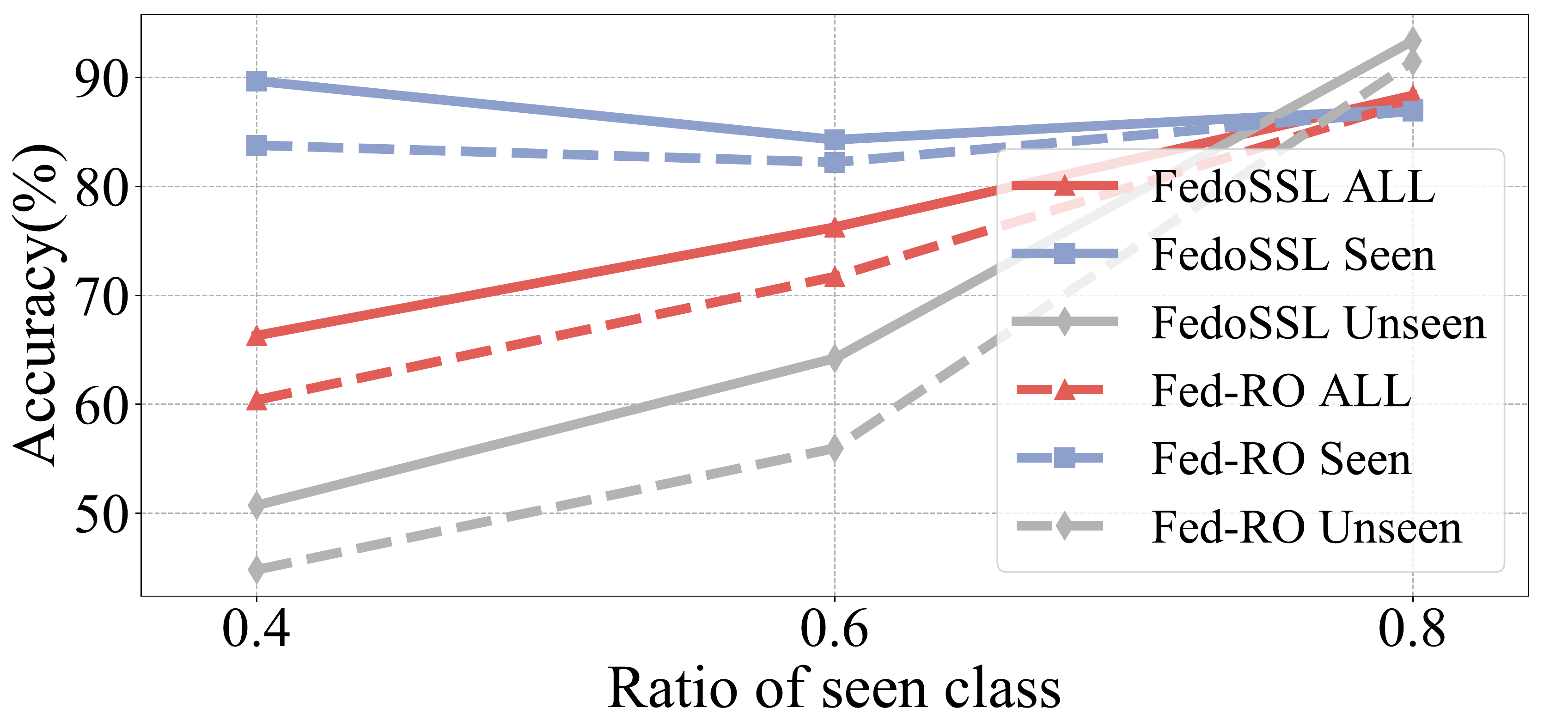}
    \caption{Performance of Fed-RO and FedoSSL with different numbers of seen classes on CIFAR-10.      
    }
    \label{fig:label_ratio}
    \vskip -0.3in
\end{figure}

\textbf{Number of Seen Class.}
To demonstrate the effectiveness of our proposed FedoSSL, we systematically evaluate the performance when changing the ratio of seen classes on CIFAR-10 dataset with 10 clients. Fed-RO is selected as the baseline because it is the best among the 4 baselines. The results in Figure~\ref{fig:label_ratio} show that FedoSSL always outperforms Fed-RO in all cases. For example, with 40\% of seen classes, FedoSSL achieves a 9.8\% overall accuracy improvement from Fed-RO. Besides, the performance of FedoSSL is stable for all ratio values and improves significantly with the increasing ratio of seen class.



\textbf{Analysis on Robustness of FedoSSL.}
To further demonstrate the robustness of our proposed FedoSSL at different experimental scales, we evaluated it in a 50-client scenario with the same other client settings, as shown in Table~\ref{tab:large_scale_fl}. The results show that FedoSSL still has the best overall accuracy in all datasets, and the performance advantage of unseen classes remains significant. 
For CIFAR-10 dataset, FedoSSL outperforms the best baselines by 6.88\% on the overall accuracy and 20.28\% on the unseen accuracy,
which is 10.74\% and 10.08\% for the CINIC-10, respectively.
Overall, our results demonstrate that FedoSSL has good robustness in large-scale FL scenarios.

\textbf{Number of Local Centroids.}
In the calibration module, the key insight is to upload local centroids instead of the representations of all data to enable computation of the global clusters. The local clustering methods can be flexibly chosen, depending on the system's constraints, e.g., for strong privacy guarantees at possibly high loss in utility, locally differentially private (local-DP) clustering methods can be used (i.e., DP-$k$-Means~\cite{balcan2017differentially}, Local-DP~\cite{chang2021locally}); for slightly weaker guarantees but higher utility, 
$K$-anonymous clustering methods can be used (i.e., $r$-Gather clustering~\cite{aggarwal2010achieving}, Sinkhorn-Knopp based clustering~\cite{genevay2019differentiable}). To further demonstrate the effectiveness of FedoSSL, we study the sensitivity of FedoSSL to the number of local centroids (i.e., $L$). Table~\ref{tab:analysis_on_number_of_cluster} shows that FedoSSL is robust to the number of local clusters as long as it even slightly exceeds the number of classes, achieving similar performance in all settings.

\begin{table}[tb]
\setlength{\abovecaptionskip}{2pt}
\caption{Sensitivity to number of local clusters on CIFAR-10. The number of global centroids is 10.}
\label{tab:analysis_on_number_of_cluster}
\begin{center}
\begin{small}
\begin{tabular}{lccccr}
\toprule
\multirow{2}{*}{$L$} &  \multirow{2}{*}{All} &  \multirow{2}{*}{Seen} & \multicolumn{3}{c}{Unseen} \\
\cmidrule(lr){4-6}
& & & LU. & GU. & AU.\\
\midrule
8  & 74.28&	84.26&	88.90&	54.09&	59.29 \\
16  & 75.76	&84.17	&89.28	&58.36	&63.15
\\
32	&76.26&	84.29	&90.68	&59.69	&64.22
\\
\bottomrule
\end{tabular}
\end{small}
\end{center}
\vskip -0.15in
\end{table}


\textbf{Analysis on Privacy.} 
Our proposed FedoSSL uses Sinkhorn-Knopp based clustering algorithm to compute $L$
equally-sized local clusters. This operation enables a 
$n/L$-anonymity privacy guarantee across all $n$ samples present on a client~\cite{lubana2022orchestra}. Furthermore, it is also feasible to use other local-DP based clustering methods instead of Sinkhorn-Knopp based clustering to provide stronger privacy guarantees. Therefore, we compare our proposed 
$K$-anonymity-guaranteed FedoSSL with the no-privacy-guaranteed version (i.e., replace Sinkhorn-Knopp based clustering with general $k$-means method). As we can see from the Table~\ref{tab:analysis_on_privacy}, FedoSSL witnesses only a small performance drop w.r.t $K$-anonymity.

\begin{table}[tb]
\setlength{\abovecaptionskip}{2pt}
\caption{Accuracy obtained using different privacy-guarnteed version of FedoSSL on CIFAR-10. `No Privacy' represents the idealized setting when local representations are shared with the server. The number of global centroids is 10.}
\label{tab:analysis_on_privacy}
\begin{center}
\begin{small}
\begin{tabular}{lccccr}
\toprule
\multirow{2}{*}{ } &  \multirow{2}{*}{All} &  \multirow{2}{*}{Seen} & \multicolumn{3}{c}{Unseen} \\
\cmidrule(lr){4-6}
& & & LU. & GU. & AU.\\
\midrule
No Privacy  & 77.19	&85.95	&89.76	&58.77	&64.05\\
$K$-anonymity	&76.26	&84.29	&90.68	&59.69	&64.22
\\
\bottomrule
\end{tabular}
\end{small}
\end{center}
\vskip -0.3in
\end{table}

\textbf{Analysis on Hyperparameters.}
We further analyze the impact of hyper-parameters $\beta$ and $\gamma$ on CINIC-10 dataset with 10 clients. Figure~\ref{subfig:a} and \ref{subfig:b} provide the performance of different $\beta$ and $\gamma$ respectively.
For $\beta$, the bigger value means the bigger effect of $\mathcal{R}_i$. When $\beta = 1$, the proposal achieves the best performance on all three accuracies and the performance does not degrade severely with $\beta$ changes. This demonstrates that FedoSSL is quite robust with the selection of $\beta$.
For $\gamma$, the bigger value means the bigger effect of $\mathcal{L}_i^{cal}$, namely, the model will be more influenced by global centroids. The results show that when $\gamma = 0.5$ the proposal achieves the best overall accuracy. When $\gamma = 0.1$, unseen classes' accuracy degrades severely, which demonstrates that the value of $\gamma$ should not be too small.

\begin{figure}[t]
\centering
\subfigure[$\beta$]{\label{subfig:a} 
 \includegraphics[width = 0.48\linewidth]{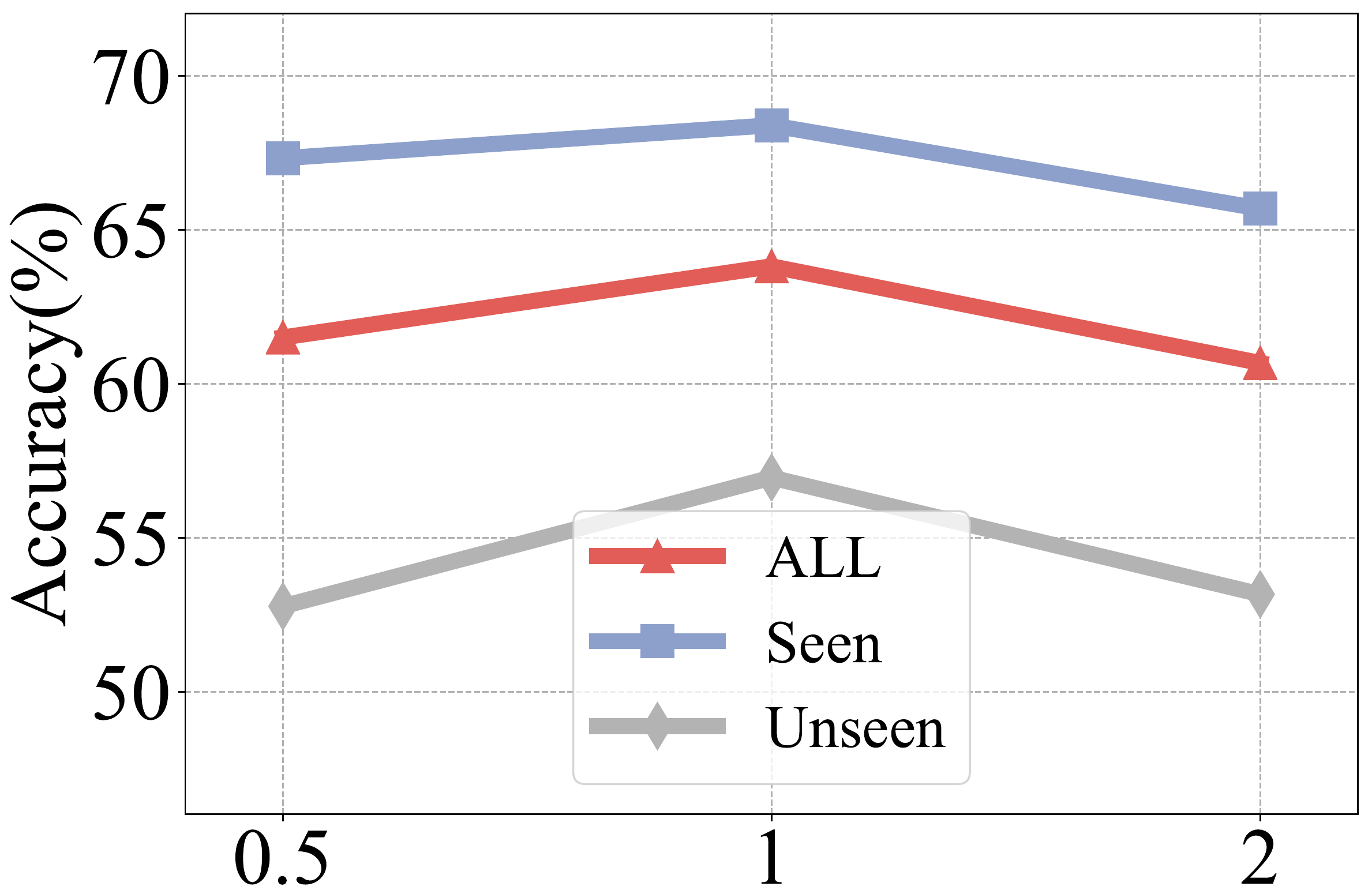}}
\subfigure[$\gamma$]{\label{subfig:b} 
 \includegraphics[width = 0.48\linewidth]{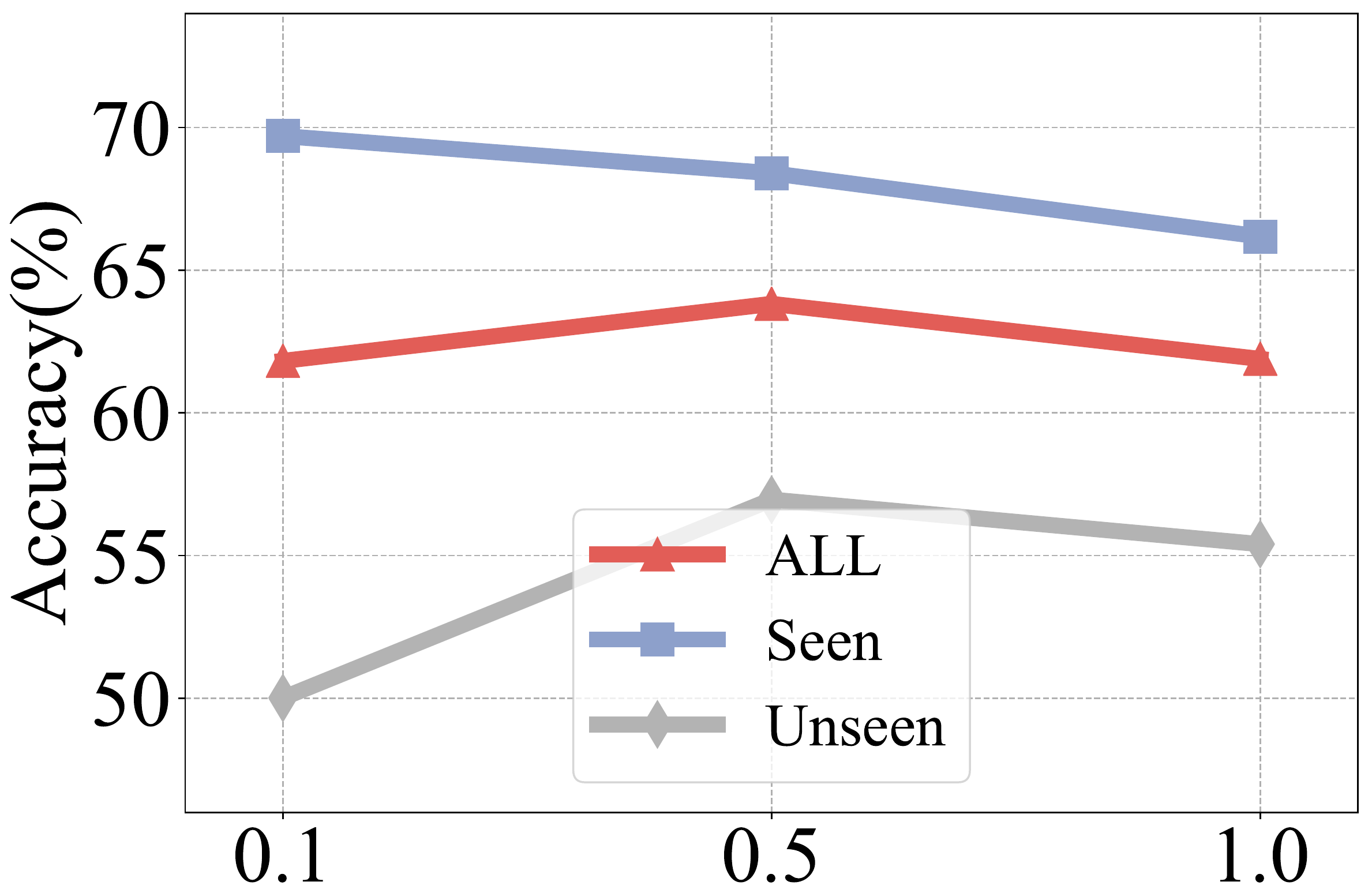}}
\caption{Classification accuracy on different 
 setting of $\beta$ (a) and $\gamma$ (b).}
\label{fig:hyperparameters}
\vspace{-0.6cm}
\end{figure}

\section{Conclusion}
In this paper, we have investigated a new federated semi-supervised learning framework under the open-world setting, 
\textbf{FedoSSL}, 
to provide unbiased training for both labeled and unlabeled data.
Specifically, we categorize the unseen classes as locally and globally unseen with the principle of whether the related unseen classes exist in only one client or not. 
An uncertainty-aware suppressed loss and a calibration module are then designed to balance the training pace among locally unseen and globally unseen classes and supplement global model aggregation phase, respectively.
%
%
We have proved that FedoSSL is a general framework for FedSSL
and can be easily compatible with state-of-the-art FL methods. 
Extensive empirical experiments have been conducted over various models and datasets to verify the effectiveness and superior performance of FedoSSL framework.

\section*{Acknowledge}
This research was supported by fundings from the Key-Area Research and Development Program of Guangdong Province (No. 2021B0101400003),  Hong Kong RGC Research Impact Fund (No. R5060-19), General Research Fund (152203/20E, 152244/21E and 152169/22E), Shenzhen Science and Technology Innovation Commission (JCYJ20200109142008673), Areas of Excellence Scheme (AoE/E-601/22-R).
\nocite{langley00}

\bibliography{example_paper}
\bibliographystyle{icml2023}

\end{document}